\documentclass[11pt,a4paper]{article}
\usepackage[hyperref]{acl2019}
\usepackage{times}
\usepackage{latexsym}
\usepackage{url}
\usepackage{graphicx}

\usepackage{xcolor}

\usepackage{natbib}

\newcommand{\myAnd}{\hspace{1em}}

\aclfinalcopy 

\title{Multilingual is not enough: BERT for Finnish}

\author{Antti Virtanen$^1$ \myAnd Jenna Kanerva \myAnd Rami Ilo$^2$ \myAnd Jouni Luoma$^3$ \myAnd Juhani Luotolahti \\[1mm] \textbf{Tapio Salakoski \myAnd Filip Ginter \myAnd Sampo Pyysalo} \\[1mm]
Turku NLP group, University of Turku \\
{\small $^1$\texttt{sajvir@utu.fi},  $^2$\texttt{rajuil@utu.fi}, $^3$\texttt{jouni.a.luoma@utu.fi}, \texttt{first.last@utu.fi}}
}

\date{}

\begin{document}
\maketitle

\begin{abstract}
Deep learning-based language models pretrained on large unannotated text corpora have been demonstrated to allow efficient transfer learning for natural language processing, with recent approaches such as the transformer-based BERT model advancing the state of the art across a variety of tasks. While most work on these models has focused on high-resource languages, in particular English, a number of recent efforts have introduced multilingual models that can be fine-tuned to address tasks in a large number of different languages. However, we still lack a thorough understanding of the capabilities of these models, in particular for lower-resourced languages. In this paper, we focus on Finnish and thoroughly evaluate the multilingual BERT model on a range of tasks, comparing it with a new Finnish BERT model trained from scratch. The new language-specific model is shown to systematically and clearly outperform the multilingual. While the multilingual model largely fails to reach the performance of previously proposed methods, the custom Finnish BERT model establishes new state-of-the-art results on all corpora for all reference tasks: part-of-speech tagging, named entity recognition, and dependency parsing. We release the model and all related resources created for this study with open licenses at \url{https://turkunlp.org/finbert} 
\end{abstract}

\section{Introduction}

Transfer learning approaches using deep neural network architectures have recently achieved substantial advances in a range of natural language processing (NLP) tasks ranging from sequence labeling tasks such as part-of-speech (POS) tagging and named entity recognition (NER) \cite{peters2018deep} to dependency parsing \cite{kondratyuk-straka-2019-75} and natural language understanding (NLU) tasks \cite{devlin2018bert}. While the great majority of this work has focused primarily on English, a number of studies have also targeted other languages, typically through multilingual models.

The BERT model of \newcite{devlin2018bert} has been particularly influential, establishing state-of-the-art results for English for a range of NLU tasks and NER when it was released. For most languages, the only currently available BERT model is the multilingual model (M-BERT) trained on pooled data from 104 languages. While M-BERT has been shown to have a remarkable ability to generalize across languages \cite{pires2019multilingual}, several studies have also demonstrated that monolingual BERT models, where available, can notably outperform M-BERT. Such results include the evaluation of the recently released French BERT model \cite{martin2019camembert}, the preliminary results accompanying the release of a German BERT model, and the evaluation of \newcite{ronnqvist-etal-2019-multilingual} comparing M-BERT with English and German monolingual models.

In this paper, we study the application of language-specific and multilingual BERT models to Finnish NLP. We introduce a new Finnish BERT model trained from scratch and perform a comprehensive evaluation comparing its performance to M-BERT on established datasets for POS tagging, NER, and dependency parsing as well as a range of diagnostic text classification tasks. The results show that 1) on most tasks the multilingual model does not represent an advance over previous state of the art, indicating that multilingual models may fail to deliver on the promise of deep transfer learning for lower-resourced languages, and 2) the custom Finnish BERT model systematically outperforms the multilingual as well as all previously proposed methods on all benchmark tasks, showing that language-specific deep transfer learning models \emph{can} provide comparable advances to those reported for much higher-resourced languages.

\section{Related Work}

The current transfer learning methods have evolved from word embedding techniques, such as word2vec \cite{mikolov2013efficient}, GLoVe \cite{pennington2014glove} and fastText \cite{joulin2016bag}, to take into account the textual context of words. Crucially, incorporating the context avoids the obvious limitations stemming from the one-vector-per-unique-word assumption inherent to the previous word embedding methods. The current successful wave of work proposing and applying different contextualized word embeddings was launched with ELMo~\cite{peters2018deep}, a context embedding method based on bidirectional LSTM networks. Another notable example is the ULMFit model \cite{howard2018universal}, which specifically focuses on techniques for domain adaptation of LSTM-based language models. Following the introduction of the attention-based (as opposed to recurrent) Transformer architecture \cite{vaswani2017attention}, BERT was proposed by \citet{devlin2018bert}, demonstrating superior performance on a broad array of tasks. The BERT model has been further refined in a number of follow-up studies \cite[e.g.][]{liu2019roberta,sanh2019distilbert} and, presently, BERT and related models form the \emph{de facto} standard approach to embedding text segments as well as individual words in context.

Unlike the previous generation of models, training BERT is a computationally intensive task, requiring substantial resources. As of this writing, Google has released English and Chinese monolingual BERT models and the multilingual M-BERT model covering 104 languages.\footnote{\url{https://github.com/google-research/bert}} Subsequently, monolingual BERT models have been published for German\footnote{\url{https://deepset.ai/german-bert}} and French \cite{martin2019camembert}. In a separate line of work, a cross-lingual BERT model for 15 languages was published by \citet{lample2019cross}, leveraging also cross-lingual signals. Finally, a number of studies have introduced monolingual models focusing on particular subdomains of English, such as BioBERT \citep{biobert} and SciBERT \citep{Beltagy2019SciBERT} for biomedical publications and scientific text.

\section{Pretraining}

We next introduce the sources of unlabeled data used to pretrain FinBERT and present the data filtering and cleanup, vocabulary generation, and pretraining processes.

\subsection{Pretraining Data}
\label{sec:pretraining-data}

\begin{table}[t!]
\centering
\small
\begin{tabular}{lrrrr}
           & Docs & Sents & Tokens & Chars \\ \hline
News       &   4M &   68M &  0.9B  & 6B  \\ 
Discussion &  83M &  351M &  4.5B  &  28B  \\ 
Crawl      &  11M &  591M &  8.1B  & 55B\\ \hline 
Total      &  98M & 1\,010M & 13.5B & 89B \\ 
\end{tabular}
\caption{Pretraining text source statistics. Tokens are counted using BERT basic tokenization. }
\label{tbl:pretraining-data-stats}
\end{table}

To provide a sufficiently large and varied unannotated corpus for pretraining, we compiled Finnish texts from three primary sources: news, online discussion, and an internet crawl. All of the unannotated texts were split into sentences, tokenized, and parsed using the Turku Neural Parser pipeline~\cite{kanerva2018turku}. Table~\ref{tbl:pretraining-data-stats} summarizes the initial statistics of the three sources prior to cleanup and filtering.

\paragraph{News} We combine two major sources of Finnish news: the Yle corpus\footnote{\url{http://urn.fi/urn:nbn:fi:lb-2017070501}}, an archive of news published by Finland's national public broadcasting company in the years 2011-2018, and The STT corpus\footnote{\url{http://urn.fi/urn:nbn:fi:lb-2019041501}} of newswire articles sent to media outlets by the Finnish News Agency (STT) between 1992 and 2018. The combined resources contain approx.\ 900 million tokens, with 20\% originating from the Yle corpus and 80\% from STT.

\paragraph{Online discussion} The Suomi24 corpus\footnote{\url{http://urn.fi/urn:nbn:fi:lb-2019010801}} (version 2017H2) contains all posts to the Suomi24 online discussion website from 2001 to 2017. Suomi24 is one of the largest social networking forums in Finland and covers a broad range of topics and levels of style and formality in language. The corpus is also roughly five times the size of the available news resources.

\begin{table}[t!]
\centering
\small
\begin{tabular}{lrrrr}
           & Docs & Sents & Tokens & Chars \\ \hline
News       &   3M &   36M &   0.5B &    4B \\ 
Discussion &  15M &  118M &   1.7B &   12B \\
Crawl      &   3M &   79M &   1.1B &    8B \\ \hline
Total      &  21M &  234M &   3.3B &   24B \\ 
\end{tabular}
\caption{Pretraining text statistics after cleanup and filtering}
\label{tbl:filtered-pretraining-data-stats}
\end{table}

\paragraph{Internet crawl} Two primary sources were used to create pretraining data from unrestricted crawls. First, we compiled documents from the dedicated internet crawl of the Finnish internet of \newcite{luotolahti2015towards} run between 2014 and 2016 using the SpiderLing crawler \cite{suchomel2012efficient}. Second, we selected texts from the Common Crawl project\footnote{\url{https://commoncrawl.org}} by running a a map-reduce language detection job on the plain text material from Common Crawl. These sources were
supplemented with plain text extracted from the Finnish Wikipedia using the \texttt{mwlib} library.
Following initial compilation, this text collection was analyzed for using the Onion deduplication tool.\footnote{\url{http://corpus.tools/wiki/Onion}} Duplicate documents were removed, and remaining documents grouped by their level of duplication.

\paragraph{Cleanup and filtering} As quality can be more important than quantity for pretraining data \cite{raffel2019exploring}, we applied a series of custom cleaning and filtering steps to the raw textual data. Initial cleaning removed header and tag material from newswire documents. In the first filtering step, machine translated and generated texts were removed using a simple support vector machine (SVM) classifier with lexical features trained on data from the FinCORE corpus \cite{laippala2019toward}. The remaining documents were then aggressively filtered using language detection and hand-written heuristics, removing documents that e.g.\ had too high a ratio of digits, uppercase or non-Finnish alphabetic characters, or had low average sentence length. A delexicalized SVM classifier operating on parse-derived features was then trained on news (positives) and heuristically filtered documents (negatives) and applied to remove documents that were morphosyntactically similar to the latter. Finally, all internet crawl-sourced documents featuring 25\% or more duplication were removed from the data.
The statistics of the final pretraining data produced in this process are summarized in Table~\ref{tbl:filtered-pretraining-data-stats}.
We note that even with this aggressive filtering, this data is roughly 30 times the size of the Finnish Wikipedia included in M-BERT pretraining data.

\begin{table}[t!]
\centering
\begin{tabular}{lllll}
                   &         & Pieces/ & UNK/   \\
Vocabulary         & Texts   & token   & token  \\ \hline
BERT cased         & En      & 1.14    & 0.0003 \\
BERT uncased       & En      & 1.10    & 0.0002 \\
M-BERT cased       & En      & 1.16    & 0.0029 \\
M-BERT uncased     & En      & 1.13    & 0.0028 \\
FinBERT cased   & Fi      & 1.43    & 0.0055 \\
FinBERT uncased & Fi      & 1.37    & 0.0022 \\
M-BERT cased       & Fi      & 1.97    & 0.0076 \\
M-BERT uncased     & Fi      & 1.86    & 0.0075 \\
\end{tabular}
\caption{Vocabulary statistics for tokenizing Wikipedia texts}
\label{tbl:vocab-eval-results}
\end{table}

\begin{table*}[t!]
\centering
\small
\begin{tabular}{l|l}
FinBERT cased    & Suomessa vaihtuu kesän aikana sekä pääministeri että valtiovarain \#\#ministeri . \\
FinBERT uncased  & suomessa vaihtuu kesan aikana seka paaministeri etta valtiovarain \#\#ministeri .\\
M-BERT cased     & Suomessa vai \#\#htuu kes \#\#än aikana sekä p \#\#ää \#\#minister \#\#i että valt \#\#io \#\#vara \#\#in \#\#minister \#\#i . \\
M-BERT uncased   & suomessa vai \#\#htuu kesan aikana seka paa \#\#minister \#\#i etta valt \#\#io \#\#vara \#\#in \#\#minister \#\#i . \\
\end{tabular}
\caption{Examples of tokenization with different vocabularies}
\label{tbl:vocab-examples}
\end{table*}

\begin{table*}[t!]
\centering
\small
\begin{tabular}{l|lll|lll|lll}
\multicolumn{1}{c}{} & \multicolumn{3}{c}{TDT} & \multicolumn{3}{c}{FTB} & \multicolumn{3}{c}{PUD} \\
          & Train  & Dev & Test & Train & Dev & Test & Train & Dev & Test \\ \hline
Sentences & 12,217 & 1,364 & 1,555   & 14,981 & 1,875  & 1,861   & --- & --- & 1,000   \\
Tokens    & 162,827 & 18,311 & 21,070  & 127,845 & 15,754 & 16,311  & --- & --- & 15,812  \\
\end{tabular}
\caption{Statistics for the Turku Dependency Treebank, FinnTreeBank and Parallel UD treebank corpora}
\label{tbl:treebank-stats}
\end{table*}

\subsection{Vocabulary generation}

To generate dedicated BERT vocabularies for Finnish, a sample of cleaned and filtered sentences were first tokenized using BERT BasicTokenizer, generating both a cased version where punctuation is separated, and an uncased version where characters are additionally mapped to lowercase and accents stripped.\footnote{We note that accent stripping makes two pairs of Finnish vowels ambiguous (a/ä and o/ö), which may be perceived as detrimental to understanding text. This step is nevertheless required for compatibility with BERT implementations.} We then used the SentencePiece \cite{kudo2018sentencepiece} implementation of byte-pair-encoding (BPE) \cite{sennrich2016neural} to generate cased and uncased vocabularies of 50,000 word pieces each.

To assess the coverage of the generated cased and uncased vocabularies and compare these to previously introduced vocabularies, we sampled a random 1\% of tokens extracted using WikiExtractor\footnote{\url{https://github.com/attardi/wikiextractor}} from the English and Finnish Wikipedias and tokenized the texts using various vocabularies to determine the number of word pieces and unknown pieces per basic token.
Table~\ref{tbl:vocab-eval-results} shows the results of this evaluation. For English, both BERT and M-BERT generate less than 1.2 WordPieces per token, meaning that the model will represent the great majority of words as a single piece. For Finnish, this ratio is nearly 2 for M-BERT. While some of this difference is explained by the morphological complexity of the language, it also reflects that only a small part of the M-BERT vocabulary is dedicated to Finnish: using the language-specific FinBERT vocabularies, this ratio remains notably lower even though the size of these vocabularies is only half of the M-BERT vocabularies.
Table~\ref{tbl:vocab-examples} shows examples of tokenization using the FinBERT and M-BERT vocabularies.

\subsection{Pretraining example generation}

We used BERT tools to create pretraining examples using the same masked language model and next sentence prediction tasks used for the original BERT. Separate duplication factors were set for news, discussion and crawl texts to create a roughly balanced number of examples from each source. We also used whole-word masking, where all pieces of a word are masked together rather than selecting masked word pieces independently. We otherwise matched the parameters and process used to create pretraining data for the original BERT, including generating separate examples with sequence lengths 128 and 512 and setting the maximum number of masked tokens per sequence separately for each (20 and 77, respectively).

\subsection{Pretraining process}

We pretrained cased and uncased models configured similarly to the base variants of BERT, with 110M parameters for each. The models were trained using 8 Nvidia V100 GPUs across 2 nodes on the Puhti supercomputer of CSC, the Finnish IT Center for Science\footnote{\url{https://research.csc.fi/csc-s-servers}}. Following the approach of \newcite{devlin2018bert}, each model was trained for 1M steps, where the initial 90\% used a maximum sequence length of 128 and the last 10\% the full 512. A batch size of 140 per GPU was used for primary training, giving a global batch size of 1120. Due to memory constraints, the batch size was dropped to 20 per GPU for training with sequence length 512. We used the LAMB optimizer \cite{you2019large} with warmup over the first 1\% of steps to a peak learning rate of 1e-4 followed by decay. Pretraining took approximately 12 days to complete per model variant.

\section{Evaluation}

We next present an evaluation of the M-BERT and FinBERT models on a series of Finnish datasets representing both downstream NLP tasks and diagnostic evaluation tasks.

Unless stated otherwise, all experiments follow the basic setup used in the experiments of \newcite{devlin2018bert}, selecting the learning rate, batch size and the number of epochs\footnote{Learning rate \{5e-5, 3e-5, 2e-5\} and epochs \{2, 3, 4\}. Batch size 32 was not used due to memory limitations.
} used for fine-tuning separately for each model and dataset combination using a grid search with evaluation on the development data. Other model and optimizer parameters were kept at the BERT defaults. Excepting for the parsing experiments, we repeat each experiment 5-10 times and report result mean and standard deviation.

\begin{table*}[t!]
\centering
\begin{tabular}{ll@{\hskip 5pt}cl@{\hskip 5pt}cl@{\hskip 5pt}c}
                            &\multicolumn{2}{c}{TDT} &\multicolumn{2}{c}{FTB} &\multicolumn{2}{c}{PUD} \\ \hline
FinBERT cased            & \textbf{98.23} & (0.04)& \textbf{98.39} & (0.03)& \textbf{98.08} &	(0.04) \\
FinBERT uncased          & 98.12	& (0.03)         & 98.28 & (0.07)         & 97.94 & (0.03) \\
M-BERT cased                & 96.97	& (0.06)         & 95.87 & (0.09)         & 97.58 & (0.03) \\
M-BERT uncased              & 96.59	& (0.05)         & 96.00 & (0.07)         & 97.48 & (0.03) \\
\cite{che2018towards}       & 97.30 & ---            & 96.70 & ---            & 97.60 & ---     \\
\cite{lim2018sex}           & 97.12 & ---            & 96.20 & ---            & 97.65 & ---     \\
\end{tabular}
\caption{Results for POS tagging (standard deviation in parentheses)}
\label{tbl:pos-results}
\end{table*}

\begin{table*}[t!]
\centering
\begin{tabular}{lrrrr}
          & Train   & Dev    & Test   & Wiki-test \\ \hline
Sentences & 13,498  &    986 &  3,512 &  3,360 \\
Tokens    & 180,178 & 13,564 & 46,363 & 49,752 \\
Entities  &  17,644 &  1,223 &  4,124 &  5,831 \\
\end{tabular}
\caption{FiNER named entity recognition corpus statistics}
\label{tbl:finer-stats}
\end{table*}

\subsection{Part of Speech Tagging}
\label{sec:part-of-speech-tagging}

Part of speech tagging is a standard sequence labeling task and several Finnish resources are available for the task.

\paragraph{Data} To assess POS tagging performance, we use the POS annotations of the three Finnish treebanks included in the Universal Dependencies (UD) collection \cite{nivre2016universal}: the Turku Dependency Treebank (TDT) \cite{pyysalo2015universal}, FinnTreeBank (FTB) \cite{voutilainen2012specifying} and Parallel UD treebank (PUD) \cite{zeman2017conll}. A broad range of methods were applied to tagging these resources as a subtask in the recent CoNLL shared tasks in 2017 and 2018 \cite{zeman2018conll}, and we use the CoNLL 2018 versions (UD version~2.2) of these corpora to assure comparability with their results. 
The statistics of these resources are shown in Table~\ref{tbl:treebank-stats}.
As the PUD corpus only provides a test set, we train and select parameters on the training and development sets of the compatibly annotated TDT corpus for evaluation on PUD.
The CoNLL shared task proceeds from raw text and thus requires sentence splitting and tokenization in order to assign POS tags. To focus on tagging performance while maintaining comparability, we predict tags for the tokens predicted by the Uppsala system \cite{smith2018universal}, distributed as part of the CoNLL'18 shared task system outputs \cite{zeman2018systemoutputs}.

\paragraph{Methods} We implement the BERT POS tagger straightforwardly by attaching a time-distributed dense output layer over the top layer of BERT and using the first piece of each wordpiece-tokenized input word to represent the word. The implementation and data processing tools are openly available.\footnote{\url{https://github.com/spyysalo/bert-pos}}
We compare POS tagging results to the best-performing methods for each corpus in the CoNLL 2018 shared task, namely that of \newcite{che2018towards} for TDT and FTB and \newcite{lim2018sex} for PUD.
We report performance for the UPOS metric as implemented by the official CoNLL 2018 evaluation script.

\paragraph{Results} Table~\ref{tbl:pos-results} summarizes the results for POS tagging. We find that neither M-BERT model improves on the previous state of the art for any of the three resources, with results ranging 0.1-0.8\% points below the best previously published results. By contrast, both language-specific models outperform the previous state of the art, with absolute improvements for FinBERT cased ranging between 0.4 and 1.7\% points. While these improvements over the already very high reference results are modest in absolute terms, the relative reductions in errors are notable: in particular, the FinBERT cased error rate on FTB is less than half of the best CoNLL'18 result \cite{che2018towards}.
We also note that the uncased models are surprisingly competitive with their cased equivalents for a task where capitalization has long been an important feature: for example, FinBERT uncased performance is within approx. 0.1\% points of FinBERT cased for all corpora.

\subsection{Named Entity Recognition}

Like POS tagging, named entity recognition is conventionally cast as a sequence labeling task. During the development of FinBERT, only one corpus was available for Finnish NER.

\paragraph{Data} FiNER, a manually annotated NER corpus for Finnish, was recently introduced by \newcite{ruokolainen2019finnish}. The corpus annotations cover five types of named entities -- person, organization, location, product and event -- as well as dates. The primary corpus texts are drawn from a Finnish technology news publication, and it additionally contains an out-of-domain test set of documents drawn from the Finnish Wikipedia.
In addition to conventional CoNLL-style named entity annotation, the corpus includes a small number of nested annotations (under 5\% of the total).
As \newcite{ruokolainen2019finnish} report results also for top-level (non-nested) annotations and the recognition of nested entity mentions would complicate evaluation, we here consider only the top-level annotations of the corpus. Table~\ref{tbl:finer-stats} summarizes the statistics of these annotations.

\begin{table*}[t!]
\centering
\begin{tabular}{ll@{\hskip 5pt}cl@{\hskip 5pt}cl@{\hskip 5pt}c}
                            & \multicolumn{2}{c}{Prec.} & \multicolumn{2}{c}{Rec.} & \multicolumn{2}{c}{F1} \\ \hline
FinBERT cased            & \textbf{91.30}& (0.12) & \textbf{93.52} & (0.10) & \textbf{92.40} & (0.09)  \\
FinBERT uncased          & 90.37 & (0.35) & 92.67 & (0.19) & 91.50 & (0.24) \\
M-BERT cased                & 89.35 & (0.21) & 91.25 & (0.17) & 90.29 & (0.14) \\
M-BERT uncased              & 88.07 & (0.25) & 90.07 & (0.22) & 89.06 & (0.21) \\
FiNER-tagger                & 90.41 & ---   & 83.51 &   ---  & 86.82 & ---    \\
\citep{gungor2018improving} & 83.59 & ---    & 85.62 &   ---  & 84.59 & ---    \\
\end{tabular}
\caption{NER results for in-domain test set (standard deviation in parentheses)}
\label{tbl:ner-results-news}
\end{table*}

\begin{table*}[t!]
\centering
\begin{tabular}{ll@{\hskip 5pt}cl@{\hskip 5pt}cl@{\hskip 5pt}c}
                            & \multicolumn{2}{c}{Prec.} & \multicolumn{2}{c}{Rec.} & \multicolumn{2}{c}{F1} \\ \hline
FinBERT cased            & 80.61 & (0.61) & \textbf{82.35} & (0.33) & \textbf{81.47} & (0.46) \\
FinBERT uncased          & 80.74 & (0.31) & 79.38 & (0.68) & 80.05 & (0.42) \\
M-BERT cased                & 75.60 & (0.49) & 76.71 & (0.61) & 76.15 & (0.50) \\
M-BERT uncased              & 75.73 & (0.73) & 71.93 & (1.01) & 73.78 & (0.81) \\
FiNER-tagger                & \textbf{88.66} &  ---  & 72.74 &  ---   & 79.91 &  ---  \\
\citep{gungor2018improving} & 67.46 &  ---   & 55.07 &  ---   & 60.64 &  ---  \\
\end{tabular}
\caption{NER results for out of domain test set (standard deviation in parentheses)}
\label{tbl:ner-results-wiki}
\end{table*}

\paragraph{Methods} Our NER implementation is based on the approach proposed for CoNLL English NER by \newcite{devlin2018bert}. A dense layer is attached on top of the BERT model to predict IOB tags independently, without a CRF layer. To include document context for each sentence, we simply concatenate as many of the following sentences as can fit in the 512 wordpiece sequence. The FiNER data does not identify document boundaries, and therefore not all these sentences are necessarily from the same document. We make the our implementation available under an open licence.\footnote{\url{https://github.com/jouniluoma/keras-bert-ner}}

We compare NER results to the rule-based FiNER-tagger~\cite{kettunen2017tagging} developed together with the FiNER corpus and to the neural network-based model of \newcite{gungor2018improving} targeted specifically toward morphologically rich languages. The former achieved the highest results on the corpus and the latter was the best-performing machine learning-based method in the experiments of \newcite{ruokolainen2019finnish}.
Named entity recognition performance is evaluated in terms of exact mention-level precision, recall and F-score as implemented by the standard \texttt{conlleval} script, and F-score is used to compare performance.

\paragraph{Results} The results for named entity recognition are summarized in Table~\ref{tbl:ner-results-news} for the in-domain (technology news) test set and Table~\ref{tbl:ner-results-wiki} for the out-of-domain (Wikipedia) test set. We find that while M-BERT is able to outperform the best previously published results on the in-domain test set, it fails to reach the performance of FiNER-tagger on the out-of-domain test set. As for POS tagging, the language-specific FinBERT model again outperforms both M-BERT as well as all previously proposed methods, establishing new state-of-the-art results for Finnish named entity recognition.

\subsection{Dependency Parsing}

Dependency parsing involves the prediction of a directed labeled graph over tokens. Finnish dependency parsing has a long history and several established resources are available for the task.

\paragraph{Data} The CoNLL 2018 shared task addressed end-to-end parsing from raw text into dependency structures on 82 different corpora representing 57 languages \cite{zeman2018conll}. We evaluate the pre-trained BERT models on the dependency parsing task using the three Finnish UD corpora introduced in Section~\ref{sec:part-of-speech-tagging}: the Turku Dependency Treebank (TDT), FinnTreeBank (FTB) and the Parallel UD treebank (PUD).
To allow direct comparison with CoNLL 2018 results, we use the same versions of the corpora as used in the shared task (UD version~2.2) and evaluate performance using the official script provided by the task organizers. These corpora are the same used in the part-of-speech tagging experiments, and their key statistics were summarized above in Table~\ref{tbl:treebank-stats}.

\begin{table*}[t!]
\centering
\begin{tabular}{l|ll|ll|ll}
\multicolumn{1}{c}{} & \multicolumn{2}{c}{TDT} & \multicolumn{2}{c}{FTB} & \multicolumn{2}{c}{PUD} \\
 Model & p.seg. & g.seg & p.seg. & g.seg. & p.seg & g.seg. \\\hline
FinBERT cased      & \textbf{91.93} & \textbf{93.56} & \textbf{92.16} & \textbf{93.95} & \textbf{92.54} & \textbf{93.10}  \\
FinBERT uncased    & 91.73 & 93.42 & 91.92 & 93.63 & 92.32 & 92.86   \\
M-BERT cased          & 86.32 & 87.99 & 85.52 & 87.46 & 89.18 & 89.75  \\
M-BERT uncased        & 86.74 & 88.61 & 86.03 & 87.98 & 89.52 & 89.95   \\
\cite{che2018towards} & 88.73 & ---   & 88.53 & ---   & 90.23 & --- \\
\cite{kulmizev2019deep} & --- & 87.0* & ---   & ---   & ---   & --- 
\end{tabular}
\caption{Labeled attachment score (LAS) parsing results for for predicted (p.seg) and gold (g.seg) segmentation. *Best performing combination in the TDT treebank (ELMo + transition-based parser).}
\label{tbl:parsing-las-results}
\end{table*}

\paragraph{Methods} We evaluate the models using the Udify dependency parser recently introduced by \citet{kondratyuk-straka-2019-75}. Udify is a multi-task model that support supporting multi- or monolingual fine-tuning of pre-trained BERT models on UD treebanks. Udify implements a multi-task network where a separate prediction layer for each task is added on top of the pre-trained BERT encoder. Additionally, instead of using only the top encoder layer representation in prediction, Udify adds a layers-wise dot-product attention, which calculates a weighted sum of all intermediate representation of 12 BERT layers for each token. All prediction layers as well as layer-wise attention are trained simultaneously, while also fine-tuning the pre-trained BERT weights.

We train separate Udify parsing models using monolingual fine-tuning for TDT and FTB. The TDT models are used to evaluate performance also on PUD, which does not include a training set. We report parser performance in terms of 
Labeled Attachment Score (LAS).
Each parser model is fine-tuned for 160 epochs with BERT weights kept frozen during the first epoch and subsequently updated along with other weights. The learning rate scheduler warm-up period is defined to be approximately one epoch. Otherwise, parameters are the same as used in \citet{kondratyuk-straka-2019-75}. As the Udify model does not implement sentence or token segmentation, we use UDPipe~\cite{udpipe:2017} to pre-segment the text when reporting LAS on predicted segmentation.

We compare our results to the best-performing system in the CoNLL 2018 shared task for the LAS metric, HIT-SCIR \cite{che2018towards}. In addition to having the highest average score over all treebanks for this metric, the system also achieved the highest LAS among 26 participants for each of the three Finnish treebanks. The dependency parser used in the HIT-SCIR system is the biaffine graph-based parser of \citet{dozat2017stanford} with deep contextualized word embeddings (ELMo)~\cite{peters-etal-2018-deep} trained monolingually on web crawl and Wikipedia data provided by \citet{conll17raw}. The final HIT-SCIR model is an ensemble over three parser models trained with different parameter initializations, where the final prediction is calculated by averaging the softmaxed output scores.

We also compare results to the recent work of \citet{kulmizev2019deep}, where the merits of two parsing architectures, graph-based \cite{kiperwasser-goldberg-2016-simple} and transition-based \cite{smith-etal-2018-82}, are studied with two different deep contextualized embeddings, ELMo and BERT. We include results for their best-performing combination on the Finnish TDT corpus, the transition-based parser with monolingual ELMo embeddings.\footnote{Note that although the UD version reported in \citet{kulmizev2019deep} is version 2.3, the results are fully comparable as there were no changes in the Finnish TDT corpus between the version 2.2 used here and version 2.3.}

\paragraph{Results} Table~\ref{tbl:parsing-las-results} shows LAS results for predicted and gold segmentation. While Udify initialized with M-BERT fails to outperform our strongest baseline \cite{che2018towards}, Udify initialized with FinBERT achieves notably higher performance on all three treebanks, establishing new state-of-the-art parsing results for Finnish with a large margin. Depending on the treebank, Udify with cased FinBERT LAS results are 2.3--3.6\% points above the previous state of the art, decreasing errors by 24\%--31\% relatively.

Casing seem to have only a moderate impact in parsing, as the performance of cased and uncased models falls within 0.1--0.6\% point range in each treebank. However, in each case the trend is that with FinBERT the cased version always outperforms the uncased one, while with M-BERT the story is opposite, the uncased always outperforming the cased one. 

To relate the high LAS of 93.56 achieved with the combination of the Udify parser and our pre-trained FinBERT model to human performance, we refer to the original annotation of the TDT corpus~\cite{haverinen2013tdt}, where individual annotators were measured against the double-annotated and resolved final annotations. The comparison is reported in terms of LAS. Here, one must take into account that the original TDT corpus was annotated in the Stanford Dependencies (SD) annotation scheme~\cite{de2008stanford}, slightly modified to be suitable for the Finnish language, while the work reported in this paper uses the UD version of the corpus. Thus, the reported numbers are not directly comparable, but keeping in mind the similarities of SD and UD annotation schemes, give a ballpark estimate of human performance in the task. \citet{haverinen2013tdt} report the average LAS of the five human annotators who participated in the treebank construction as 91.3, with individual LAS scores ranging from 95.9 to 71.8 (or 88.0 ignoring an annotator who only annotated 2\% of the treebank and was still in the training phrase). Based on these numbers, the achieved parser LAS of 93.56 seems to be on par with or even above average human level performance and approaching the level of a well-trained and skilled annotator.

\begin{table*}[t!]
\centering
\begin{tabular}{ll@{\hskip 5pt}ll@{\hskip 5pt}ll@{\hskip 5pt}ll@{\hskip 5pt}ll@{\hskip 5pt}l}
        & \multicolumn{2}{c}{1K} & \multicolumn{2}{c}{$\sim$3K} & \multicolumn{2}{c}{10K} & \multicolumn{2}{c}{$\sim$32K} & \multicolumn{2}{c}{100K} \\ \hline
FinBERT cased	& 87.99	& (0.35)	& 89.49	& (0.11)	& 90.57	& (0.15)	& \textbf{91.42} & (0.14)	& 91.74	& (0.13) \\
FinBERT uncased	& \textbf{87.86} & (0.37) & \textbf{89.52} & (0.13)	& \textbf{90.58} & (0.09)	& 91.23	& (0.08)	& \textbf{91.76} & (0.10) \\
M-BERT  cased	    & 83.22	& (0.72)	& 86.56	& (0.18)	& 88.44	& (0.14)	& 89.34	& (0.22)	& 90.28	& (0.18) \\
M-BERT  uncased	    & 84.92	& (0.37)	& 87.14	& (0.26)	& 88.69	& (0.15)	& 89.63	& (0.11)	& 90.49	& (0.19) \\
FastText            & 78.50 & (0.00)    & 81.71 & (0.03)    & 85.90 & (0.00)    & 88.36 & (0.05)    & 89.40 & (0.00) \\
\end{tabular}
\caption{Yle news 10-class text classification accuracy for varying training set sizes (percentages, standard deviation in parentheses)}
\label{tbl:full-results-yle}
\end{table*}

\begin{table*}[t!]
\centering
\begin{tabular}{ll@{\hskip 5pt}ll@{\hskip 5pt}ll@{\hskip 5pt}ll@{\hskip 5pt}ll@{\hskip 5pt}l}
        & \multicolumn{2}{c}{1K} & \multicolumn{2}{c}{$\sim$3K} & \multicolumn{2}{c}{10K} & \multicolumn{2}{c}{$\sim$32K} & \multicolumn{2}{c}{100K} \\ \hline
FinBERT cased	& 75.00	& (0.34)	& 77.48	& (0.17)	& 79.18	& (0.20)	& 80.89	& (0.16)	& 82.51	& (0.12) \\
FinBERT uncased	& \textbf{75.71} & (0.24) & \textbf{77.88} & (0.24) & \textbf{79.79} & (0.20)	& \textbf{81.25} & (0.12)	& \textbf{82.80} & (0.14) \\
M-BERT cased	    & 45.28	& (12.65)	& 59.09	& (2.72)	& 67.92	& (0.43)	& 72.84	& (0.15)	& 76.51	& (0.16) \\
M-BERT uncased	    & 51.20	& (3.76)	& 63.13	& (0.42)	& 69.01	& (0.35)	& 73.89	& (0.29)	& 77.38	& (0.19) \\
FastText            & 47.74 & (0.05)    & 56.66 & (0.05)    & 64.27 & (0.05)    & 70.86 & (0.05)    & 74.71 & (0.03) \\
\end{tabular}
\caption{Ylilauta online discussion 10-class text classification accuracy for varying training set sizes (percentages, standard deviation in parentheses)}
\label{tbl:full-results-ylilauta}
\end{table*}

\begin{figure*}[t!]
\centering
\includegraphics[width=\textwidth]{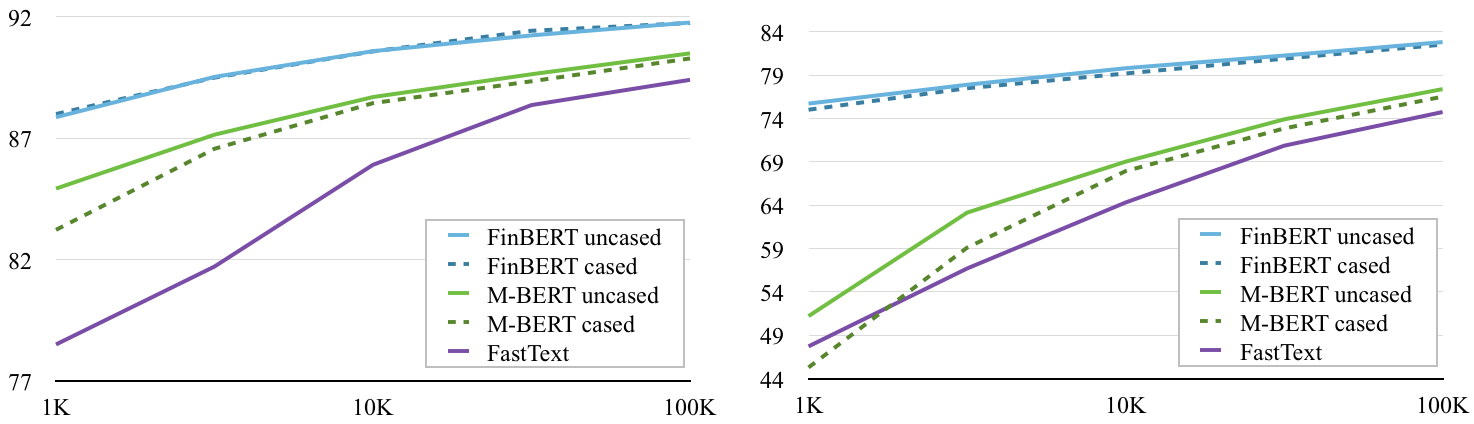}
\caption{Text classification accuracy with different training data sizes for Yle news (left) and Ylilauta online discussion (right). (Note log $x$ scales and different $y$ ranges.)}
\label{fig:yle-ylilauta-curves}
\end{figure*}

\subsection{Text classification}

Finnish lacks the annotated language resources to construct a comprehensive collection of classification tasks such as those available for English \cite{rajpurkar2016squad,wang2018glue,zellers2018swag}. To assess model performance at text classification, we create two datasets based on Finnish document collections with topic information, one representing formal language (news) and the other informal (online discussion).

\paragraph{Data} Documents in the Yle news corpus (Section~\ref{sec:pretraining-data}) are annotated using a controlled vocabulary to identify subjects such as \emph{sports}, \emph{politics}, and \emph{economy}. We identified ten such upper-level topics that were largely non-overlapping in the data and sampled documents annotated with exactly one selected topic to create a ten-class classification dataset. As the Yle corpus is available for download under a license that does not allow redistribution, we release tools to recreate this dataset.\footnote{\url{https://github.com/spyysalo/yle-corpus}}
The Ylilauta corpus\footnote{\url{http://urn.fi/urn:nbn:fi:lb-2015031802}} consists of the text of discussions on the Finnish online discussion forum Ylilauta from 2012 to 2014. Each posted message belongs to exactly one board, with topics such as \emph{games}, \emph{fashion} and \emph{television}. We identified the ten most frequent topics and sampled messages consisting of at least ten tokens to create a text classification dataset from the Ylilauta data.\footnote{\url{https://github.com/spyysalo/ylilauta-corpus}}

To facilitate analysis and comparison, we downsample both corpora to create balanced datasets with 10000 training examples as well as 1000 development and 1000 test examples of each class. To reflect generalization performance to new documents, both resources were split chronologically, drawing the training set from the oldest texts, the test set from the newest, and the development set from texts published between the two. To assess classifier performance across a range of training dataset sizes, we further downsampled the training sets to create versions with 100, 316, 1000, and 3162 examples of each class ($10^2, 10^{2.5}, \ldots$). Finally, we truncated each document to a maximum of 256 basic tokens to minimize any advantage the language-specific model might have due to its more compact representation of Finnish.

\paragraph{Methods} We implement the text classification methods following \newcite{devlin2018bert}, minimizing task-specific architecture and simply attaching a dense output layer to the initial (\texttt{[CLS]}) token of the top layer of BERT. We establish baseline text classification performance using fastText\footnote{\url{https://fasttext.cc/}} \cite{joulin2016bag}. We evaluated a range of parameter combinations and different pretrained word vectors for the method using the development data, selecting character n-gram features of lengths 3--7, training for 25 epochs, and initialization with subword-enriched embeddings induced from Wikipedia texts\footnote{\url{https://fasttext.cc/docs/en/pretrained-vectors.html}} \cite{bojanowski2017enriching} for the final experiments.

\paragraph{Results} The text classification results for various training set sizes are shown in Table~\ref{tbl:full-results-yle} for Yle news and in Table~\ref{tbl:full-results-ylilauta} for Ylilauta online discussion and illustrated in Figure~\ref{fig:yle-ylilauta-curves}.
We first note that performance is notably higher for the news corpus, with error rates for a given method and data set size more than doubling when moving from news to the discussion corpus. As both datasets represent 10-class classification tasks with balanced classes, this suggests that the latter task is inherently more difficult, perhaps in part due to the incidence of spam and off-topic messages on online discussion boards.

The cased and uncased variants of FinBERT perform very similarly for both datasets and all training set sizes, while for M-BERT the uncased model consistently outperforms the cased -- as was also found for parsing -- with a marked advantage for small dataset sizes.

\begin{table*}[t!]
\centering
\begin{tabular}{lllllllll}
\multicolumn{1}{c}{} & \multicolumn{2}{c}{FinBERT cased} & \multicolumn{2}{c}{FinBERT uncased} & \multicolumn{2}{c}{M-BERT cased} & \multicolumn{2}{c}{M-BERT uncased} \\ \hline
BiShift &   \textbf{72.10} & (1.72) & 70.75 & (1.08) & 62.73 & (0.85) & 62.17 & (1.36) \\
CoordInv &  78.29 & (3.30) & \textbf{79.57} & (2.32) & 76.00 & (1.82) & 78.93 & (1.63) \\
ObjNum &    \textbf{78.34} & (0.53) & 75.80 & (3.76) & 74.64 & (0.89) & 75.38 & (0.86) \\
Tense &     96.26 & (0.19) & \textbf{96.49} & (0.51) & 95.12 & (0.08) & 96.32 & (0.20) \\
SentLen &   \textbf{40.41} & (0.97) & 39.70 & (0.43) & 38.35 & (0.63) & 39.95 & (0.63) \\
SubjNum &   83.81 & (0.70) & \textbf{85.72} & (0.29) & 83.22 & (0.48) & 84.86 & (0.40) \\
TreeDepth & 38.57 & (1.07) & 38.65 & (0.81) & \textbf{38.75} & (0.25) & 38.58 & (0.89) \\
WC &        \textbf{11.05} & (0.34) &  9.70 & (0.30) & 10.33 & (0.55) & 10.98 & (0.38) \\
\end{tabular}
\caption{Probing results (standard deviation in parentheses).}
\label{tbl:probing-results}
\end{table*}

Comparing M-BERT and FinBERT, we find that the language-specific models outperform the multilingual models across the full range of training data sizes for both datasets. For news, the four BERT variants have broadly similar learning curves, with the absolute advantage for FinBERT models ranging from 3\% points for 1K examples to just over 1\% point for 100K examples, and relative reductions in error from 20\% to 13\%. For online discussion, the differences are much more pronounced, with M-BERT models performing closer to the FastText baseline than to FinBERT. Here the language-specific BERT outperforms the multilingual by over 20\% points for the smallest training data and maintains a 5\% point absolute advantage even with 100,000 training examples, halving the error rate of the multilingual model for the smallest training set and maintaining an over 20\% relative reduction for the largest.

These contrasting results for the news and discussion corpora may be explained in part by domain mismatch: while the news texts are written in formal Finnish resembling the Wikipedia texts included as pretraining data for all BERT models as well as the FastText word vectors, only FinBERT pretraining material included informal Finnish from online discussions.\footnote{The online discussions included in FinBERT pretraining data were drawn from the Suomi24 corpus and thus did not include any of the Ylilauta messages used in this evaluation.} This suggests that in pretraining BERT models care should be taken to assure that not only the targeted language but also the targeted text domains are sufficiently represented in the data.

\subsection{Probing Tasks}

Finally, we explored the ability of the models to capture linguistic properties using the probing tasks proposed by \citet{Conneau_2018}.
We use the implementation and Finnish data introduced for these tasks by \citet{ravishankar-etal-2019-multilingual},\footnote{\url{https://github.com/ltgoslo/xprobe}} which omit the \textbf{TopConst} task defined in the original paper. We also left out the Semantic odd-man-out (SOMO) task, as we found the data to have errors making the task impossible to perform correctly.
All of the tasks involve freezing the BERT layers and training a dense layer on top of it to function as a diagnostic classifier. The only information passed from BERT to the classifier is the state represented by the \texttt{[CLS]} token.

In brief, the tasks can be roughly categorized into 3 different groups: surface, syntactic and semantic information.

\paragraph{Surface tasks}

In the \textit{sentence length} (\textbf{SentLen}) task, sentences are classified into 6 classes depending on their length.
The \textit{word content} (\textbf{WC}) task measures the model's ability to determine which of 1000 mid-frequency words occurs in a sentence, where only one of the words is present in any one sentence.

\paragraph{Syntactic tasks}

The \textit{tree depth} (\textbf{TreeDepth}) task is used to test how well the model can identify the depth of the syntax tree of a sentence. We used dependency trees to maintain comparability with the work of \citet{ravishankar-etal-2019-multilingual}, whereas the original task used constituency trees.
\textit{Bigram shift} (\textbf{BiShift}) tests the model's ability to recognize when two adjacent words have had their positions swapped.

\paragraph{Semantic tasks}

In the \textit{subject number} (\textbf{SubjNum}) task the number of the subject, i.e.\  singular or plural, connected to the main verb of a sentence is predicted. 
\textit{Object number} (\textbf{ObjNum}) is similar to the previous task but for objects of the main verb.
The \textit{Coordination inversion} (\textbf{CoordInv}) has the order of two clauses joined by a coordinating conjunction reversed in half the examples. The model then has to predict whether or not a given example was inverted.
In the \textbf{Tense} task the classifier has to predict whether a main verb of a sentence is in the present or past tense.

\paragraph{Results} Table~\ref{tbl:probing-results} presents results comparing the FinBERT models to replicated M-BERT results from
\citet{ravishankar-etal-2019-multilingual}. We find that the best performance is achieved by either the cased or uncased language-specific model for all tasks except TreeDepth, where M-BERT reaches the highest performance. The differences between the results for the language-specific and multilingual models are modest for most tasks with the exception of the BiShift task, where the FinBERT models are shown to be markedly better at identifying sentences with inverted words. While this result supports the conclusion of our other experiments that FinBERT is the superior language model, results for the other tasks offer only weak support at best. We leave for future work the question whether these tasks measure aspects where the language-specific model does not have a clear advantage over the multilingual or if the results reflect limitations in the implementation or data of the probing tasks.

\section{Discussion}

We have demonstrated that it is possible to create a language-specific BERT model for a lower-resourced language, Finnish,  that clearly outperforms the multilingual BERT at a range of tasks and advances the state of the art in many NLP tasks. These findings raise the question whether it would be possible to realize similar advantages for other languages that currently lack dedicated models of this type. It is likely that the feasibility of training high quality deep transfer learning models hinges on the availability of pretraining data.

As of this writing, Finnish ranks 24th among the different language editions of Wikipedia by article count,\footnote{\url{https://en.wikipedia.org/wiki/List_of_Wikipedias}} and 25th in Common Crawl by page count.\footnote{\url{https://commoncrawl.github.io/cc-crawl-statistics/plots/languages}} There are thus dozens of languages for which unannotated corpora of broadly comparable size or larger than that used to pretrain FinBERT could be readily assembled from online resources.
Given that language-specific BERT models have been shown to outperform multilingual ones also for high-resource languages such as French \cite{martin2019camembert} -- ranked 3rd by Wikipedia article count -- it is further likely that the benefits of a language-specific model observed here extend at least to languages with more resources than Finnish.
(We are not aware of efforts to establish the minimum amount of unannotated text required to train high-quality models of this type.)

The methods we applied to collect and filter texts for training FinBERT have only few language dependencies, such as the use of UD parsing results for filtering. As UD resources are already available for over 70 languages, the specific approach and tools introduced in this work could be readily applied to a large number of languages. To facilitate such efforts, we also make all of the supporting tools developed in this work available under open licenses.

\section{Conclusions}

In this work, we compiled and carefully filtered a large unannotated corpus of Finnish, trained language-specific FinBERT models, and presented evaluations comparing these to multilingual BERT models at a broad range of natural language processing tasks. The results indicate that the multilingual models fail to deliver on the promises of deep transfer learning for lower-resourced languages, falling behind the performance of previously proposed methods for most tasks. By contrast, the newly introduced FinBERT model was shown not only to outperform multilingual BERT for all downstream tasks, but also to establish new state-of-the art results for three different Finnish corpora for part-of-speech tagging and dependency parsing as well as for named entity recognition.

The FinBERT models and all of the tools and resources introduced in this paper are available under open licenses from 
\url{https://turkunlp.org/finbert}.

\section*{Acknowledgments}

We gratefully acknowledge the support of CSC – IT Center for Science through its Grand Challenge program, the Academy of Finland, the Google Digital News Innovation Fund and collaboration of the Finnish News Agency STT, as well as the NVIDIA Corporation GPU Grant Program.  \\

\bibliographystyle{acl_natbib}
\bibliography{main}

\end{document}